\documentclass{article}
\usepackage{ijcai15}

\usepackage{times}
\usepackage{balance}
\usepackage{epsfig}
\usepackage{graphicx}
\usepackage{amsmath}
\usepackage{amssymb}
\usepackage{bm}
\usepackage{algorithm}
\usepackage{algpseudocode}
\usepackage[tight,footnotesize]{subfigure}

\title{Salient Object Detection via Augmented Hypotheses}
\author{Tam V. Nguyen \and Jose Sepulveda\\
Department for Technology, Innovation and Enterprise\\
Singapore Polytechnic\\
\{nguyen\_van\_tam, sepulveda\_jose\}@sp.edu.sg}

\begin{document}

\maketitle

\begin{abstract}
In this paper, we propose using \textit{augmented hypotheses} which consider objectness, foreground and compactness for salient object detection. Our algorithm consists of four basic steps. First, our method generates the objectness map via objectness hypotheses. Based on the objectness map, we estimate the foreground margin and compute the corresponding foreground map which prefers the foreground objects. From the objectness map and the foreground map, the compactness map is formed to favor the compact objects. We then derive a saliency measure that produces a pixel-accurate saliency map which uniformly covers the objects of interest and consistently separates fore- and background. We finally evaluate the proposed framework on two challenging datasets, MSRA-1000 and iCoSeg. Our extensive experimental results show that our method outperforms state-of-the-art approaches.
\end{abstract}

\section{Introduction}
The ultimate goal of salient object detection is to search for salient objects which draw human attention on the image. The research has shown that computational models simulating low-level stimuli-driven attention \cite{Koch,IT} are quite successful and represent useful tools in many practical scenarios, including image resizing~\cite{FT}, attention retargeting~\cite{Attention}, dynamic captioning~\cite{TamMM13}, image classification~\cite{Qiang} and action recognition~\cite{STAP}. The existing methods  can be classified into biologically-inspired and computationally-oriented approaches. On the one hand, works belonging to the first class \cite{IT,RC} are generally based on the model proposed by Koch and Ullman~\cite{Koch}, in which the low-level stage processes features such as color, orientation of edges, or direction of movement. One example of this model is the work by Itti et al. \cite{IT}, which use a Difference of Gaussians approach to evaluate those features. However, the resulting saliency maps are generally blurry, and often overemphasize small, purely local features, which renders this approach less useful for applications such as segmentation, detection, etc \cite{RC}.

\begin{figure}[!t]
\centering
\includegraphics[width = \linewidth]{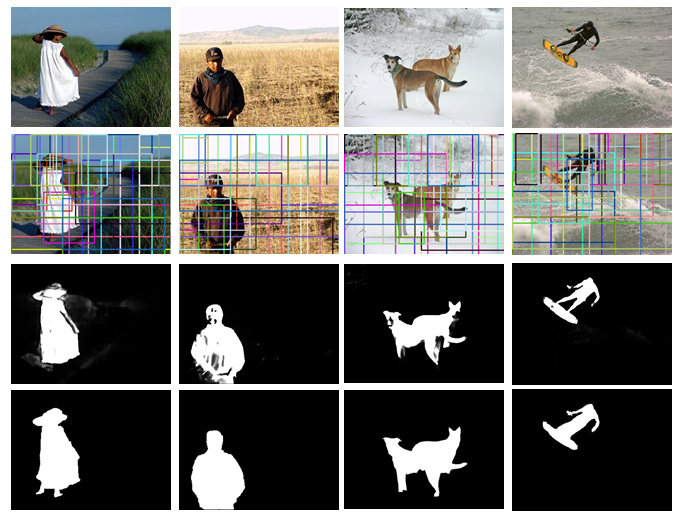}
\caption{From top to bottom: original images, the objectness hypotheses, results of our saliency computation, and ground truth labeling. For a better viewing, only 40 object hypotheses are displayed in each image.}
\label{fig:introduction}
\end{figure}

\begin{figure*}[!t]
\centering
\includegraphics[width = \linewidth]{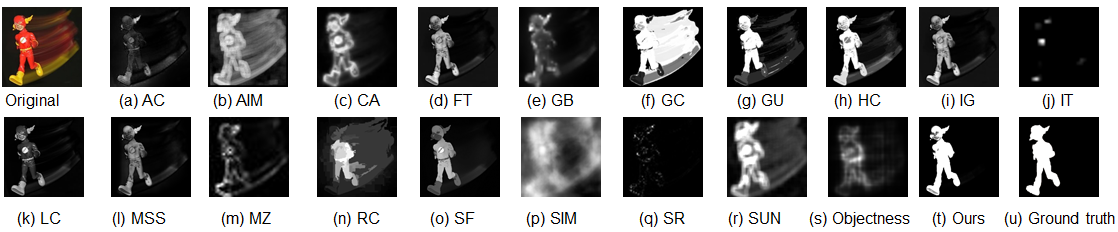}
\caption{Saliency maps computed by our proposed AH method (t) and state-of-the-art methods (a-r), salient region detection (AC~\protect\cite{AC}), attention based on information maximization (AIM~\protect\cite{AIM}), context-aware (CA~\protect\cite{CA}), frequency-tuned (FT~\protect\cite{FT}), graph based saliency (GB~\protect\cite{GBVS}), global components (GC~\protect\cite{GC}), global uniqueness (GU~\protect\cite{GC}), global contrast saliency (HC and RC~\protect\cite{RC}), spatial temporal cues (LC~\protect\cite{LC}), visual attention measurement (IT~\protect\cite{IT}), maximum symmetric surround (MSS~\protect\cite{MSS}), fuzzy growing (MZ~\protect\cite{MZ}), saliency filters (SF~\protect\cite{SF}), induction model (SIM~\protect\cite{SIM}), spectral residual (SR~\protect\cite{SR}), saliency using natural statistics (SUN~\protect\cite{SUN}), and the objectness map (s). Our result (t) focuses on the main salient object as shown in ground truth (u).}
\label{fig:comparison}
\end{figure*} 

On the other hand, computational methods relate to typical applications in computer vision and graphics. For example, frequency space methods \cite{SR} determine saliency based on spectral residual of the Fourier transform of an image. The resulting saliency maps exhibit undesirable blurriness and tend to highlight object boundaries rather than its entire area. Since human vision is sensitive to color, different approaches use local or global analysis of color contrast. Local methods estimate the saliency of a particular image region based on immediate image neighborhoods, e.g., based on dissimilarities at the pixel-level \cite{MZ} or histogram analysis~\cite{RC}. While such approaches are able to produce less blurry saliency maps, they are agnostic of global relations and structures, and they may also be more sensitive to high frequency content like image edges and noise. In a global manner, \cite{FT} achieves globally consistent results by computing color dissimilarities to the mean image color. Murray et al. \cite{SIM} introduced an efficient model of color appearance, which contains a principled selection of parameters as well as an innate spatial pooling mechanism. There also exist different patch-based methods which estimate dissimilarity between image patches \cite{CA,SF}. While these algorithms are more consistent in terms of global image structures, they suffer from the involved combinatorial complexity, hence they are applicable only to relatively low resolution images, or they need to operate in spaces of reduced image dimensionality \cite{AIM}, resulting in loss of salient details.

\begin{figure*}[!t]
\centering
\subfigure[Original image ]
{
\includegraphics[width = 0.126\linewidth]{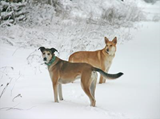}
}
\subfigure[Hypotheses]
{
\includegraphics[width = 0.126\linewidth]{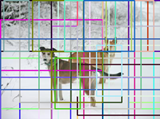}
}
\subfigure[Objectness ]
{
\includegraphics[width = 0.126\linewidth]{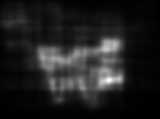}
}
\subfigure[Margin]
{
\includegraphics[width = 0.126\linewidth]{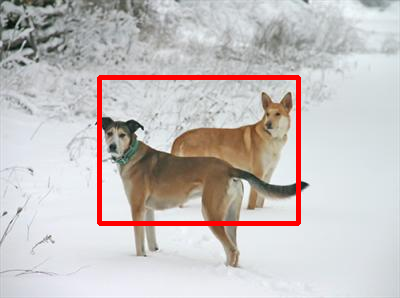}
}
\subfigure[Foreground]
{
\includegraphics[width = 0.126\linewidth]{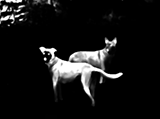}
}
\subfigure[Compactness ]
{
\includegraphics[width = 0.126\linewidth]{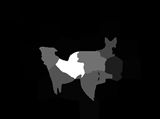}
}
\subfigure[Saliency map]
{
\includegraphics[width = 0.126\linewidth]{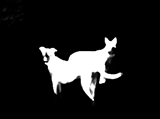}
}
\caption{Illustration of the main phases of our algorithm. The object hypotheses are generated from the input image. The objectness map is later formed by accumulating all hypotheses. The foreground map is then created from the difference between the pixel's color and the  background color obtained following the estimated margins. We then oversegment the image into superpixels and compute the compactness map based on the spatial distribution of superpixels. Finally, a saliency value is assigned to each pixel.}
\label{fig:step}
\end{figure*}

Despite many recent improvements, the difficult question is still whether ``the salient object is a real object''. That question bridges the problem of salient object detection into the traditional object detection research. In the latter object detection problem, the efficient sliding window object detection while keeping the computational cost feasible is very important. Therefore, there exist numerous objectness hypothesis generation methods proposing a small number (e.g. 1,000) of category-independent hypotheses, that are expected to cover all objects in an image~\cite{Lampert,Alexe,Uijlings,BING}. Objectness hypothesis is usually represented as a value which reflects how likely an image window covers an object of any category. Lampert et al.~\cite{Lampert} introduced a branch-and-bound scheme for detection. However, it can only be used to speed up classifiers that users can provide a good bound on highest score. Alexe et al. \cite{Alexe} proposed a cue integration approach to get better prediction performance more efficiently. Uijlings et al.~\cite{Uijlings} proposed a selective search approach to get higher prediction performance. However, these methods are time-consuming, taking ~$3$ seconds for one image. Recently, Cheng et al. \cite{BING} presented a simple and fast objectness measure by using binarized normed gradients features which compute the objectness of each image window at any scale and aspect ratio only requires a few bit operations. This method can be run 1,000+ times faster than popular alternatives.

In this work, we investigate applying objectness to the problem of salient object detection. We utilize the object hypotheses from the objectness hypothesis generation augmented with foreground and compactness constraint in order to produce a fast and high quality salient object detector. The exemplary object hypotheses and our saliency prediction are shown in the second and the third row of Figure \ref{fig:introduction}, respectively. As we demonstrate in our experimental evaluation, each of our individual measures already performs close to or even better than some existing approaches, and our combined method currently achieves the best ranking results on two public datasets provided by \cite{FT,icoseg}. Figure \ref{fig:comparison} shows the comparison of our saliency map to other baselines in literature. The main contributions of this work can be summarized as follows. 
\begin{itemize}
\item We conduct the comprehensive study on how the objectness hypotheses affect the salient object detection.
\item We propose the foreground map and compactness map, derived from the objectness map, which can cover both global and local information of the saliency object.
\item Unlike other works in the literature, we evaluate our proposed method on two challenging datasets in order to know the impact of our work in different settings.
\end{itemize}

\begin{figure}[!b]
\centering
\includegraphics[width = \linewidth]{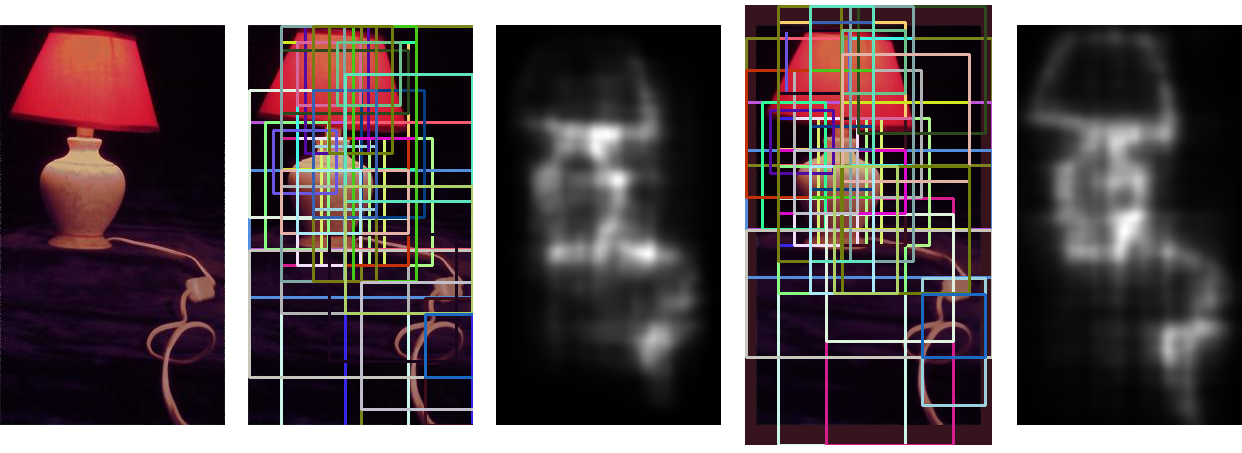}
\caption{From left to right: the original image, the object hypotheses and the corresponding objectness map, the extended object hypotheses and the corresponding objectness map.}
\label{fig:extension}
\end{figure}

\section{Methodology}

In this section, we describe the details of our augmented hypotheses (AH), and we show how the objectness measures as well as the saliency assignment can be efficiently computed. Figure \ref{fig:step} illustrates the overview of our processing steps.

\subsection{Objectness Map}
In this work, we extract object hypotheses from the input image to form the objectness map. We assume that the salient objects attract more object hypotheses than other parts in the image. As aforementioned, the objectness hypothesis generators propose a small number $n_p$ (e.g. 1,000) of category-independent hypotheses, that are expected to cover all objects in an image. Each hypothesis $P_i$ has coordinate $(l_i, t_i, r_i, b_i)$, where $l_i, t_i$ are the coordinate of the top left point, whereas $r_i, b_i$ are the coordinate of the bottom right point. Here, we formulate each hypothesis $P_i$ $\in \mathbb{R}^{H \times W}$, where $H$ and $W$ are the height and the width of the input image $\bm{I}$, respectively. The value of each element $P_i(x,y)$ is defined as:

\begin{equation}
P_i(x,y) = \begin{cases}
    1 \quad \phantom{\infty}\text{if}\,\, t_i \leq x \leq b_i \,\,\text{and}\,\, l_i \leq y \leq r_i   \\
    0 \quad \phantom{\infty} otherwise
      \end{cases}.
\end{equation} 
The objectness map is constructed by accumulating all object hypotheses:
\begin{equation}
OB(x,y) = \sum_{i=1}^{n_p}P_i(x,y).
\end{equation}

The objectness map is later rescaled into the range [0..1]. We observe that the objectness map discourages the object parts locating close to the image boundary. Thus we extend the original image by embedding an image border with the size is $10\%$ of the original image's size. The addition image border is filled with the mean color of the original image. We perform the hypothesis extraction and compute the objectness map similar to the aforementioned steps. The final objectness map is cropped to the size of the original image. Figure \ref{fig:extension} demonstrates the effect of our image extension and the shrinkage of the objectness map. 

\subsection{Foreground Map}
The salient object tends to be distinctive from its surrounding context. Thus, we aim to model the background which can facilitate the object localization. In particular, the foreground map is computed by finding the difference between the color of the original image and the background image. In order to model the background, we first localize the salient object by the margin shown as the red rectangle in Fig \ref{fig:step}d. To this end, we compute the accumulate objectness level by four directions $n_r$, namely, top, bottom, left, and right. For each direction, the accumulated objectness level is bounded by a threshold  $\theta$. To boost this process, we utilize the integral image \cite{Viola} computed from the objectness map. Finally, there are $n_r$, 4 in this work, corresponding rectangles surrounding the salient object. Each bounding rectangle $r_i$ is represented by  its mean color $\mu_{r_i}$. The foreground value computed for each pixel $(x,y)$ is computed as follows,
 
\begin{equation}
FG(x,y) = \prod_{i=1}^{n_r}\|\bm{I}(x,y) - \bm{\mu}_{r_i}\|,
\end{equation}
where $\bm{I}(x,y)$ is the color vector of the pixel $(x,y)$. 

\subsection{Compactness Map}
The foreground map prefers the color of the salient object of the foreground. Unfortunately, it also favors the similar color appearing in the background. We observe that though the colors belonging to the background will be distributed over the entire image exhibiting a high spatial variance, the foreground objects are generally more compact~\cite{SF}. Therefore, we compute the compactness map in order to remove the noise from the background. First, we compute the centroid of interest $(x_c, y_c) = (\frac{\sum_{(x,y)}x \times OF(x,y)}{\sum_{(x,y)}OF(x,y)}, \frac{\sum_{(x,y)}y \times OF(x,y)}{\sum_{(x,y)}OF(x,y)})$, where the objectness-foreground value $OF(x,y) = OB(x,y) \times FG(x,y)$. Intuitively, the pixel close to the centroid of interest tends to be more salient, whereas the farther pixels tend to be less salient. In addition, the saliency value of a certain pixel reduces if the path between the centroid and that pixel contains many low saliency values. The naive method is to compute the path from the centroid of interest to other pixels. However, it is time-consuming to perform this task in the pixel-level. Therefore, we transform it to superpixel-level. The image is over-segmented into superpixels, and the $OF$ value of a superpixel is computed as the average $OF$ values of all containing pixels. The over-segmented image can be formulated as a graph $\mathbb{G} = (\mathbb{V},\mathbb{E})$, where $\mathbb{V}$ is the list of vertices (superpixels) and $\mathbb{E}$ is the list of edges connecting the neighboring superpixels. 

\begin{algorithm}[!t]
  \caption{Superpixel compactness computation}\label{alg:1}
  \begin{algorithmic}[1]
	\State $\bm{l} = \{v_c\}$.
	\State $\bm{c} = \bm{0}\in\mathbb{R}^{n_{sp}}$.
	\State $\bm{t} = \emptyset$
	\While {$l \neq \emptyset$}
		\For{each vertex $\bm{v}_i$ in $\bm{l}$}
			\For{each edge $(\bm{v}_i, \bm{v}_j)$}
				\If{$c(v_j) < \sqrt{c(v_i) \times OF(v_j)}$ }
					\State 	$c(v_j) \leftarrow \sqrt{c(v_i) \times OF(v_j)}$
			       	\State $\bm{t}\leftarrow \bm{t}\bigcup \bm{v}_j$;
				\EndIf	
			\EndFor
		\EndFor
		\State $\bm{l}\leftarrow \bm{t}$
		\State $\bm{t} = \emptyset$
	\EndWhile
	\State \textbf{return} compactness values $\bm{c}$ of superpixels.
  \end{algorithmic}
\end{algorithm}

The procedure to compute the compactness values of superpixels is summarized in Algorithm \ref{alg:1}. Denote $\bm{v}_c$ as the superpixel containing the centroid of interest. The algorithm transfers the $OF$ value from the $\bm{v}_c$ to all other superpixels. The procedure performs a sequence of relaxation steps, namely assigning the compactness value $\bm{c}(\bm{v}_j)$ of superpixel $\bm{v}_j$ by the square root of its neighboring superpixel's compactness value and its own $OF$ value. Our algorithm only relaxes edges from vertices $\bm{v}_j$ for which $\bm{c}(\bm{v}_j)$ has recently changed, since other vertices cannot lead to correct relaxations. Additionally, the algorithm may be terminated early when no recent changes exist. Finally, the compactness value $CN$ is computed as:
\begin{equation}
CN(x,y) = \bm{c}(sp(x,y)),
\end{equation}
where $sp(x,y)$ returns the index of the superpixel containing pixel $(x,y)$.
\subsection{Saliency Assignment}
We normalize the objectness map $OB$, foreground map $FG$, and compactness map $CN$ to the range [0..1]. We assume that all measures are independent, and hence we combine these terms as follows to compute a saliency value $S$ for each pixel:

\begin{equation}
S(x,y) = OB(x,y) \times FG(x,y) \times CN(x,y).
\end{equation}

The resulting pixel-level saliency map may have an arbitrary scale. In the final step, we rescale the saliency values within [0..1] and to contain at least 10\% saliency pixels.

\begin{figure*}[!t]
\centering
\subfigure[Fixed threshold]{
\includegraphics[width = 0.32\linewidth]{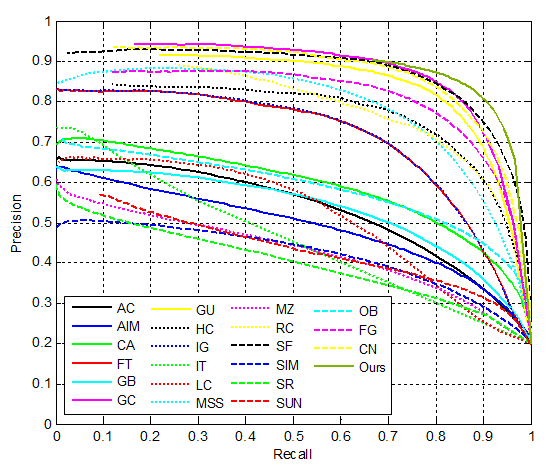}
	}
\subfigure[Adaptive threshold]{
\includegraphics[width = 0.32\linewidth]{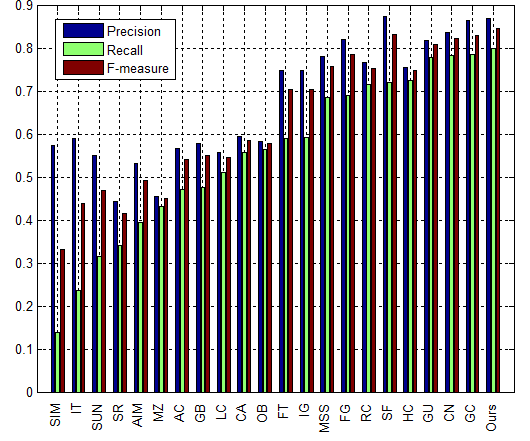}
}
\subfigure[Mean absolute error]{
\includegraphics[width = 0.32\linewidth]{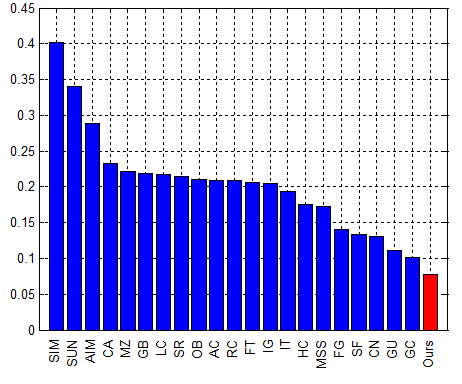}
}
\caption{Statistical comparison with 18 saliency detection methods using all the 1000 images from MSRA-1000 dataset \protect\cite{FT} with pixel accuracy saliency region annotation: (a) the average precision recall curve by segmenting saliency maps using fixed thresholds, (b) the average precision recall by adaptive thresholding (using the same method as in FT~\protect\cite{FT}, SF~\protect\cite{SF}, GC~\protect\cite{GC}, etc.), (c) the mean absolute error of the different saliency methods to ground truth mask. Please check Figure~\ref{fig:comparison} for the references to the publications in which the baseline methods are presented. }
\label{fig:res_MSRA}
\end{figure*}

\begin{figure*}[!t]
\centering
\includegraphics[width = \linewidth]{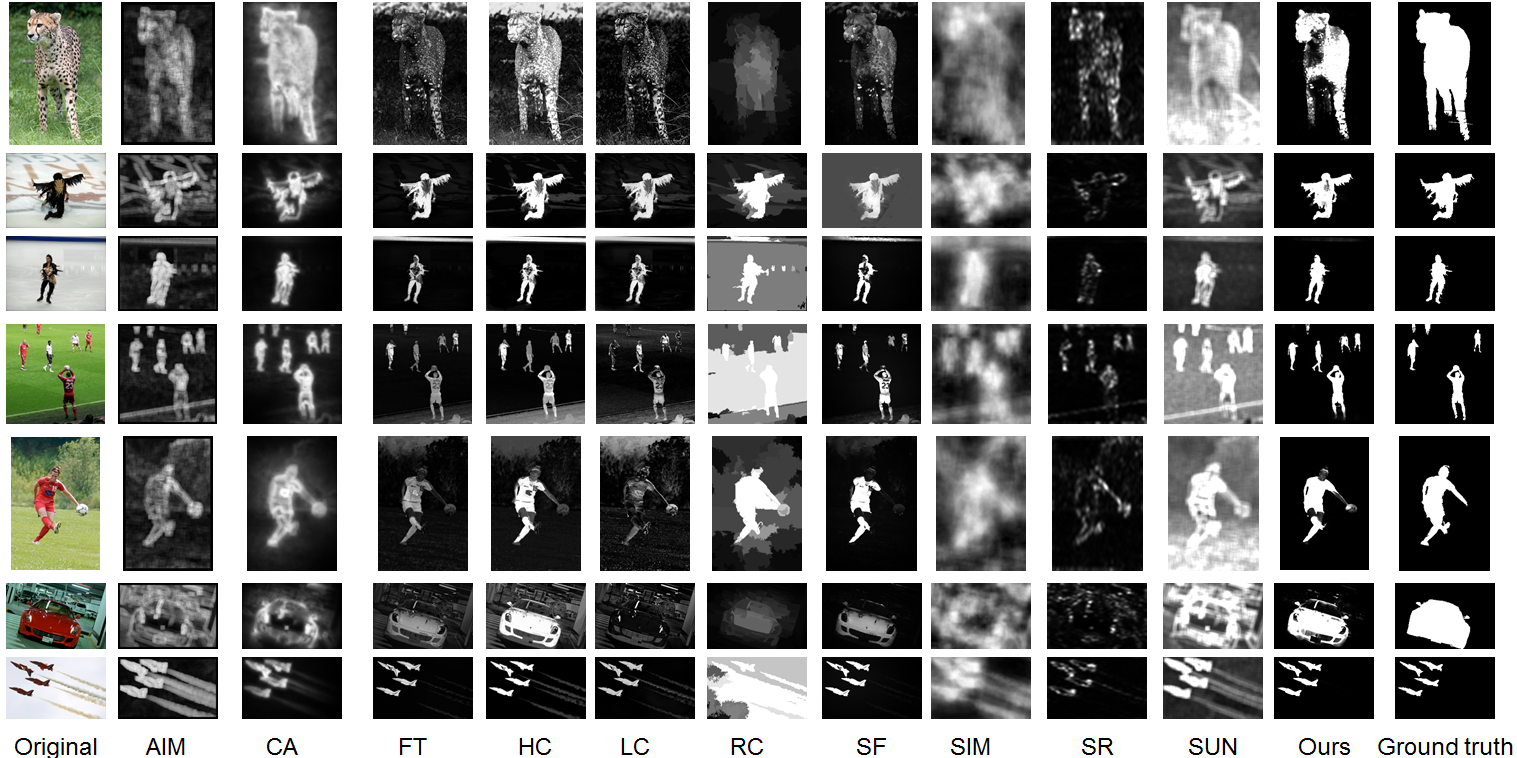}
\caption{Visual comparison of saliency maps on iCoSeg dataset. We compare our method (AH) to other 10 alternative methods. Our results are close to ground truth and focus on the main salient objects.}
\label{fig:results}
\end{figure*}

\subsection{Implementation Settings}
We apply the state-of-the-art objectness detection technique, i.e., binarized normed gradients (BING)~\cite{BING}, to produce a set of candidate object windows. Our selection of BING is two-fold. First, BING extractor has a weak training from the simple feature, e.g., binarized normed gradients. Therefore, it is useful comparing to bottom-up edge extractor. Second, the BING extractor is able to run 10 times faster than real-time, i.e., 300 frames per second (fps). BING hypothesis generator is trained with VOC2007 dataset \cite{VOC} same as in \cite{BING}. In order to compute the foreground map, $\theta$ is set as $0.1$ and we convert the color channels from RGB to Lab color space as suggested in \cite{FT,SF}. Regarding the image over-segmentation, we use SLIC~\cite{SLIC} for the superpixel segmentation. We set the number of superpixels as $100$ as a trade-off between the fine over-segmentation and the processing time. 
\section{Evaluation}
\subsection{Datasets and Evaluation Metrics}
We evaluate and compare the performances of our algorithm against previous baseline algorithms on two representative benchmark datasets: the \textbf{MSRA 1000} salient object dataset \cite{FT} and the Interactive cosegmentation Dataset (\textbf{iCoSeg}) \cite{icoseg}. The MSRA-1000 dataset contains 1,000 images with the pixel-wise ground truth provided by \cite{FT}. Note that each image in this dataset contains a salient object. Meanwhile, the iCoSeg contains 643 images with single or multiple objects in a single image. 

The first evaluation compares the precision and recall rates. High recall can be achieved at the expense of reducing the precision and vice-versa so it is important to evaluate
both measures together. In the first setting, we compare binary masks for every threshold in the range [0..255]. In the second setting, we use the image dependent adaptive threshold proposed by~\cite{FT}, defined as twice the mean saliency of the image:

\begin{equation}
T_a = \frac{2}{W \times H} \sum_{(x,y)}S(x,y). 
\end{equation}

In addition to precision and recall we compute their weighted harmonic mean measure or $F-measure$,
which is defined as:
\begin{equation}
F_{\beta} = \frac{(1 + \beta^2)\times Precision \times Recall}{\beta^2 \times Precision + Recall}.
\end{equation}
As in previous methods \cite{FT,GC,SF}, we use $\beta^2$ = 0.3.

For the second evaluation, we follow Perazzi et al. \cite{SF} to evaluate the mean absolute error (MAE) between a continuous saliency map S and the binary ground truth G
for all image pixels $(x,y)$, defined as:
\begin{equation}
MAE = \frac{1}{W \times H}\sum_{(x,y)}{|S(x,y) - G(x,y)|}.
\end{equation}

\begin{figure*}[!t]
\centering
\subfigure[Fixed threshold]{
\includegraphics[width = 0.32\linewidth]{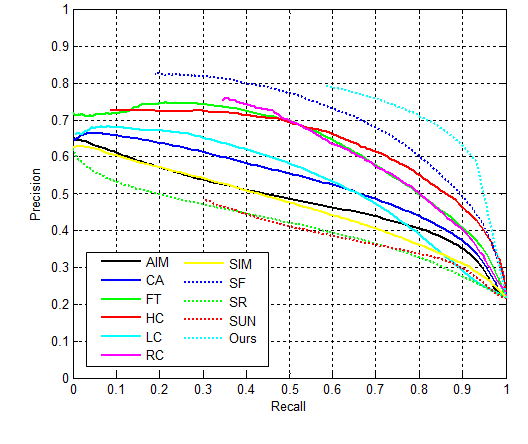}
}
\subfigure[Adaptive threshold]{
\includegraphics[width = 0.32\linewidth]{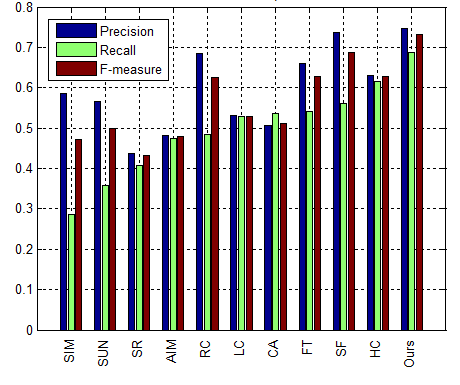}
}
\subfigure[Mean absolute error]{
\includegraphics[width = 0.32\linewidth]{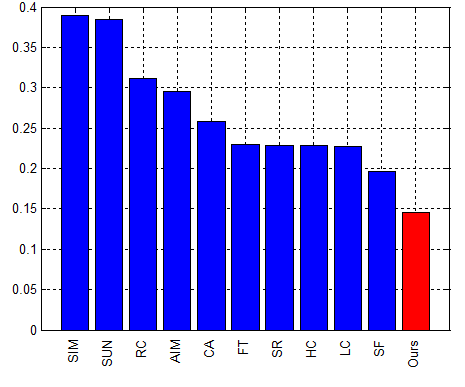}
}
\caption{Statistical comparison with 10 saliency detection methods using all the 643 images from iCoSeg benchmark \protect\cite{icoseg} with pixel accuracy saliency region annotation: (a) the average precision recall curve by segmenting saliency maps using fixed thresholds, (b) the average precision recall by adaptive thresholding (using the same method as in FT~\protect\cite{FT}, GC~\protect\cite{GC}, etc.), (c) the mean absolute error of the different saliency methods to ground truth mask.}
\label{fig:res_Icoseg}
\end{figure*}

\subsection{Performance on MSRA1000 dataset}

Following~\cite{FT,SF,GC}, we first evaluate our methods using a precision/recall curve which is shown in Figure~\ref{fig:res_MSRA}. Our work reaches the highest precision/recall rate over all baselines. As a result, our method also obtains the best performance in terms of F-measure. We also evaluate the individual components in our system, namely, objectness map (OB), foreground map (FG), and compactness map (CN). They generally achieve the acceptable performance which is comparable to other baselines. The performance of the objectness map itself is outperformed by our proposed augmented hypotheses.  In this work, our novelty is that we adopt and augment the conventional hypotheses by adding two key features: foregroundness and compactness to detect salient objects. When fusing them together, our unified system achieves the state-of-the-art performance in every single evaluation metric. 

As discussed in the SF~\cite{SF} and GC~\cite{GC}, neither the precision nor recall measure considers the true negative counts. These measures favor methods which successfully assign saliency to salient pixels but fail to detect non-salient regions over methods that successfully do the opposite. Instead, they suggested that MAE is a better metric than precision recall analysis for this problem. As shown in Figure \ref{fig:res_MSRA}c, our work outperforms the state-of-the-art performance \cite{GC} by 24\%. One may argue that a simple boosting of saliency values similar as in ~\cite{SF} results would improve it. However, a boosting of saliency values could easily result in the boosting of low saliency values related to background that we also aim to avoid. 

\subsection{Performance on iCoSeg dataset}
The iCoSeg dataset is ``less popular'' in the sense that some baselines do not even release detection results and source-code. We only reproduced 10 methods on iCoSeg thanks to their existing source-code.
The visual comparison of saliency maps generated from our method and different baselines are demonstrated in Figure \ref{fig:results}. Our results are close to ground truth and focus on the main salient objects. We first evaluate our methods using a precision/recall curve which is shown in Figure~\ref{fig:res_Icoseg}a, b. Our method outperforms all other baselines in both two settings, namely fixed threshold and adaptive threshold. As shown in Figure \ref{fig:res_Icoseg}c, our method achieves the best performance in terms of MAE. Our work outperforms other methods by a large margin, 25\%. 

\subsection{Computational Efficiency}
It is also worth investigating the computational efficiency of different methods. In Table \ref{table:times}, we compare the average running time of our approach to the currently best performing methods on the benchmark images. We compare the performance of our method in terms of speed with methods with most competitive accuracy (GC~\cite{GC}, SF~\cite{SF}).  The average time of each method is measured on a PC with Intel i7 $3.3$ GHz CPU and $8$GB RAM. Performance of all the methods compared in this table are based on implementations in C++ and MATLAB. The CA method the slowest one because it requires an exhaustive nearest-neighbor search among patches. Meanwhile, our method is able to run in a real-time manner. Our procedure spends most of the computation time on generating the objectness map (about 35\%) and forming the compactness map (about 50\%). From the experimental results, we find that our algorithm is effective and computationally efficient. 

\begin{table}[!t]
\caption{Comparison of running times in the MSRA 1000 benchmark~\protect\cite{FT}.}
\begin{tabular}{|l|c|c|c|c|c|}
\hline
Method~~~~   & ~~CA~~     & ~~RC~~   & ~~SF~~   & ~~GC~~   & ~~Ours~~ \\ \hline
Time (s)~~~~ & 51.2   & 0.14 & 0.15 & 0.09 & 0.07 \\ \hline
Code     & Matlab & C++  & C++  & C++  & C++  \\ \hline
\end{tabular}
\label{table:times}
\end{table}

\section{Conclusion and Future Work}
In this paper, we have presented a novel method, \textit{augmented hypotheses} (AH), which adopts the object hypotheses in order to rapidly detect \textit{salient objects}. To this end, three maps are derived from object hypotheses: superimposed hypotheses form an objectness map, a foreground map is computed from deviations in color from the background, and a compactness map emerges from propagating saliency labels in the oversegmented image. These three maps are fused together to detect salient objects with sharp boundaries.  Experimental results on two challenging datasets show that our results are 24\% - 25\% better than the previous best results (compared against 10+ methods in two different datasets), in terms of mean absolute error while also being faster. 

For future work, we aim to investigate more sophisticated techniques for objectness measures and integrate more cues, i.e., depth ~\cite{Lang} and audio~\cite{Yanxiang} information. Also, we would like to study the impact of salient object detection into the object hypothesis process.  

\section{Acknowledgments}
This work was supported by Singapore Ministry of Education under research Grants MOE2012-TIF-2-G-016 and MOE2014-TIF-1-G-007. 

\bibliographystyle{named}
\bibliography{egbib}

\begin{thebibliography}{}

\bibitem[\protect\citeauthoryear{Achanta and S{\"{u}}sstrunk}{2010}]{MSS}
Radhakrishna Achanta and Sabine S{\"{u}}sstrunk.
\newblock Saliency detection using maximum symmetric surround.
\newblock In {\em ICIP}, pages 2653--2656, 2010.

\bibitem[\protect\citeauthoryear{Achanta \bgroup \em et al.\egroup }{2008}]{AC}
Radhakrishna Achanta, Francisco~J. Estrada, Patricia Wils, and Sabine
  S{\"{u}}sstrunk.
\newblock Salient region detection and segmentation.
\newblock In {\em International Conference of Computer Vision Systems}, pages
  66--75, 2008.

\bibitem[\protect\citeauthoryear{Achanta \bgroup \em et al.\egroup }{2009}]{FT}
Radhakrishna Achanta, Sheila~S. Hemami, Francisco~J. Estrada, and Sabine
  S{\"{u}}sstrunk.
\newblock Frequency-tuned salient region detection.
\newblock In {\em CVPR}, pages 1597--1604, 2009.

\bibitem[\protect\citeauthoryear{Achanta \bgroup \em et al.\egroup
  }{2012}]{SLIC}
Radhakrishna Achanta, Appu Shaji, Kevin Smith, Aur{\'{e}}lien Lucchi, Pascal
  Fua, and Sabine S{\"{u}}sstrunk.
\newblock {SLIC} superpixels compared to state-of-the-art superpixel methods.
\newblock {\em T-PAMI}, 34(11):2274--2282, 2012.

\bibitem[\protect\citeauthoryear{Alexe \bgroup \em et al.\egroup
  }{2012}]{Alexe}
Bogdan Alexe, Thomas Deselaers, and Vittorio Ferrari.
\newblock Measuring the objectness of image windows.
\newblock {\em T-PAMI}, 34(11):2189--2202, 2012.

\bibitem[\protect\citeauthoryear{Batra \bgroup \em et al.\egroup
  }{2010}]{icoseg}
Dhruv Batra, Adarsh Kowdle, Devi Parikh, Jiebo Luo, and Tsuhan Chen.
\newblock icoseg: Interactive co-segmentation with intelligent scribble
  guidance.
\newblock In {\em CVPR}, pages 3169--3176, 2010.

\bibitem[\protect\citeauthoryear{Bruce and Tsotsos}{2005}]{AIM}
Neil Bruce and John Tsotsos.
\newblock Saliency based on information maximization.
\newblock In {\em NIPS}, 2005.

\bibitem[\protect\citeauthoryear{Chen \bgroup \em et al.\egroup }{2012}]{Qiang}
Qiang Chen, Zheng Song, Yang Hua, ZhongYang Huang, and Shuicheng Yan.
\newblock Hierarchical matching with side information for image classification.
\newblock In {\em CVPR}, pages 3426--3433, 2012.

\bibitem[\protect\citeauthoryear{Chen \bgroup \em et al.\egroup
  }{2014}]{Yanxiang}
Yanxiang Chen, Tam~V. Nguyen, Mohan~S. Kankanhalli, Jun Yuan, Shuicheng Yan,
  and Meng Wang.
\newblock Audio matters in visual attention.
\newblock {\em T-CSVT}, 24(11):1992--2003, 2014.

\bibitem[\protect\citeauthoryear{Cheng \bgroup \em et al.\egroup }{2011}]{RC}
Ming{-}Ming Cheng, Guo{-}Xin Zhang, Niloy~J. Mitra, Xiaolei Huang, and
  Shi{-}Min Hu.
\newblock Global contrast based salient region detection.
\newblock In {\em CVPR}, pages 409--416, 2011.

\bibitem[\protect\citeauthoryear{Cheng \bgroup \em et al.\egroup }{2013}]{GC}
Ming{-}Ming Cheng, Jonathan Warrell, Wen{-}Yan Lin, Shuai Zheng, Vibhav Vineet,
  and Nigel Crook.
\newblock Efficient salient region detection with soft image abstraction.
\newblock In {\em CVPR}, pages 1529--1536, 2013.

\bibitem[\protect\citeauthoryear{Cheng \bgroup \em et al.\egroup }{2014}]{BING}
Ming-Ming Cheng, Ziming Zhang, Wen-Yan Lin, and Philip H.~S. Torr.
\newblock {BING}: Binarized normed gradients for objectness estimation at
  300fps.
\newblock In {\em CVPR}, 2014.

\bibitem[\protect\citeauthoryear{Everingham \bgroup \em et al.\egroup
  }{2010}]{VOC}
Mark Everingham, Luc~Van Gool, Christopher Williams, John Winn, and Andrew
  Zisserman.
\newblock The pascal visual object classes {(VOC)} challenge.
\newblock {\em IJCV}, 88(2):303--338, 2010.

\bibitem[\protect\citeauthoryear{Goferman \bgroup \em et al.\egroup
  }{2010}]{CA}
Stas Goferman, Lihi Zelnik{-}Manor, and Ayellet Tal.
\newblock Context-aware saliency detection.
\newblock In {\em CVPR}, pages 2376--2383, 2010.

\bibitem[\protect\citeauthoryear{Harel \bgroup \em et al.\egroup }{2006}]{GBVS}
Jonathan Harel, Christof Koch, and Pietro Perona.
\newblock Graph-based visual saliency.
\newblock In {\em NIPS}, pages 545--552, 2006.

\bibitem[\protect\citeauthoryear{Hou and Zhang}{2007}]{SR}
Xiaodi Hou and Liqing Zhang.
\newblock Saliency detection: {A} spectral residual approach.
\newblock In {\em CVPR}, 2007.

\bibitem[\protect\citeauthoryear{Itti \bgroup \em et al.\egroup }{1998}]{IT}
Laurent Itti, Christof Koch, and Ernst Niebur.
\newblock A model of saliency-based visual attention for rapid scene analysis.
\newblock {\em T-PAMI}, 20(11):1254--1259, 1998.

\bibitem[\protect\citeauthoryear{Koch and Ullman}{1985}]{Koch}
C~Koch and S~Ullman.
\newblock {Shifts in selective visual attention: towards the underlying neural
  circuitry.}
\newblock {\em Hum Neurobiol}, 1985.

\bibitem[\protect\citeauthoryear{Lampert \bgroup \em et al.\egroup
  }{2008}]{Lampert}
Christoph~H. Lampert, Matthew~B. Blaschko, and Thomas Hofmann.
\newblock Beyond sliding windows: Object localization by efficient subwindow
  search.
\newblock In {\em CVPR}, 2008.

\bibitem[\protect\citeauthoryear{Lang \bgroup \em et al.\egroup }{2012}]{Lang}
Congyan Lang, Tam~V. Nguyen, Harish Katti, Karthik Yadati, Mohan~S.
  Kankanhalli, and Shuicheng Yan.
\newblock Depth matters: Influence of depth cues on visual saliency.
\newblock In {\em ECCV}, pages 101--115, 2012.

\bibitem[\protect\citeauthoryear{Ma and Zhang}{2003}]{MZ}
Yu{-}Fei Ma and HongJiang Zhang.
\newblock Contrast-based image attention analysis by using fuzzy growing.
\newblock In {\em ACM MM}, pages 374--381, 2003.

\bibitem[\protect\citeauthoryear{Murray \bgroup \em et al.\egroup }{2011}]{SIM}
Naila Murray, Maria Vanrell, Xavier Otazu, and C.~Alejandro P{\'{a}}rraga.
\newblock Saliency estimation using a non-parametric low-level vision model.
\newblock In {\em CVPR}, pages 433--440, 2011.

\bibitem[\protect\citeauthoryear{Nguyen \bgroup \em et al.\egroup
  }{2013a}]{Attention}
Tam~V. Nguyen, Bingbing Ni, Hairong Liu, Wei Xia, Jiebo Luo, Mohan Kankanhalli,
  and Shuicheng Yan.
\newblock Image re-attentionizing.
\newblock {\em Multimedia, IEEE Transactions on}, 15(8):1910--1919, 2013.

\bibitem[\protect\citeauthoryear{Nguyen \bgroup \em et al.\egroup
  }{2013b}]{TamMM13}
Tam~V. Nguyen, Mengdi Xu, Guangyu Gao, Mohan Kankanhalli, Qi~Tian, and
  Shuicheng Yan.
\newblock Static saliency vs. dynamic saliency: a comparative study.
\newblock In {\em ACM MM}, pages 987--996, 2013.

\bibitem[\protect\citeauthoryear{Nguyen \bgroup \em et al.\egroup
  }{2015}]{STAP}
Tam~V. Nguyen, Zheng Song, and Shuicheng Yan.
\newblock {STAP}: Spatial-temporal attention-aware pooling for action
  recognition.
\newblock {\em T-CSVT}, 2015.

\bibitem[\protect\citeauthoryear{Perazzi \bgroup \em et al.\egroup }{2012}]{SF}
Federico Perazzi, Philipp Kr{\"{a}}henb{\"{u}}hl, Yael Pritch, and Alexander
  Hornung.
\newblock Saliency filters: Contrast based filtering for salient region
  detection.
\newblock In {\em CVPR}, pages 733--740, 2012.

\bibitem[\protect\citeauthoryear{Uijlings \bgroup \em et al.\egroup
  }{2013}]{Uijlings}
Jasper Uijlings, Koen van~de Sande, Theo Gevers, and Arnold Smeulders.
\newblock Selective search for object recognition.
\newblock {\em IJCV}, 104(2):154--171, 2013.

\bibitem[\protect\citeauthoryear{Viola and Jones}{2001}]{Viola}
Paul~A. Viola and Michael~J. Jones.
\newblock Robust real-time face detection.
\newblock In {\em {ICCV}}, page 747, 2001.

\bibitem[\protect\citeauthoryear{Zhai and Shah}{2006}]{LC}
Yun Zhai and Mubarak Shah.
\newblock Visual attention detection in video sequences using spatiotemporal
  cues.
\newblock In {\em ACM MM}, pages 815--824, 2006.

\bibitem[\protect\citeauthoryear{Zhang \bgroup \em et al.\egroup }{2008}]{SUN}
Lingyun Zhang, Matthew~H. Tong, Tim~K. Marks, Honghao Shan, and Garrison~W.
  Cottrell.
\newblock Sun: A bayesian framework for saliency using natural statistics.
\newblock {\em Journal of Vision}, 8(7), 2008.

\end{thebibliography}

\end{document}